%% file: ArXiv.tex
\documentclass[10pt,twocolumn,letterpaper]{article}

\usepackage[pagenumbers]{cvpr} 

\usepackage{graphicx}
\usepackage{amsmath}
\usepackage{amssymb}
\usepackage{booktabs}
\usepackage{xcolor}

\definecolor{limegreen}{rgb}{0.2, 0.8, 0.2}
\definecolor{ered}{rgb}{0.72, 0.16, 0.2}

\usepackage[pagebackref,breaklinks,colorlinks]{hyperref}

\newcommand{\VL}{V\&L {}}
\newcommand{\CLIPI}{h^{\mathcal{I}}}
\newcommand{\CLIPT}{h^{\mathcal{T}}}

\usepackage[capitalize]{cleveref}
\crefname{section}{Sec.}{Secs.}
\Crefname{section}{Section}{Sections}
\Crefname{table}{Table}{Tables}
\crefname{table}{Tab.}{Tabs.}

\begin{document}

\title{Improving Zero-Shot Models with Label Distribution Priors}

\author{Jonathan Kahana  $\quad$  Niv Cohen $\quad$ Yedid Hoshen\\
School of Computer Science and Engineering\\
The Hebrew University of Jerusalem, Israel\\ 
{\tt\small jonathan.kahana@mail.huji.ac.il}\\ \\ 
Project webpage: \textcolor{blue}{https://www.vision.huji.ac.il/clippr}} 
\maketitle

\begin{abstract}

Labeling large image datasets with attributes such as facial age or object type is tedious and sometimes infeasible. Supervised machine learning methods provide a highly accurate solution, but require manual labels which are often unavailable. Zero-shot models (e.g., CLIP) do not require manual labels but are not as accurate as supervised ones, particularly when the attribute is numeric. We propose a new approach, CLIPPR\footnote{The code is available under: \textcolor{blue}{https://github.com/jonkahana/CLIPPR}} (\textbf{CLIP} with \textbf{Pr}iors), which adapts zero-shot models for regression and classification on unlabelled datasets. Our method does not use any annotated images. Instead, we assume a prior over the label distribution in the dataset. We then train an adapter network on top of CLIP under two competing objectives: i) minimal change of predictions from the original CLIP model ii) minimal distance between predicted and prior distribution of labels. Additionally, we present a novel approach for selecting prompts for Vision \& Language models using a distributional prior. Our method is effective and presents a significant improvement over the original model. We demonstrate an improvement of $28\%$ in mean absolute error on the UTK age regression task. We also present promising results for classification benchmarks, improving the classification accuracy on the ImageNet dataset by $2.83\%$, without using any labels.

\end{abstract}


\section{Introduction}

\begin{figure*}[t!]
    \centering
    \includegraphics[width=14cm]{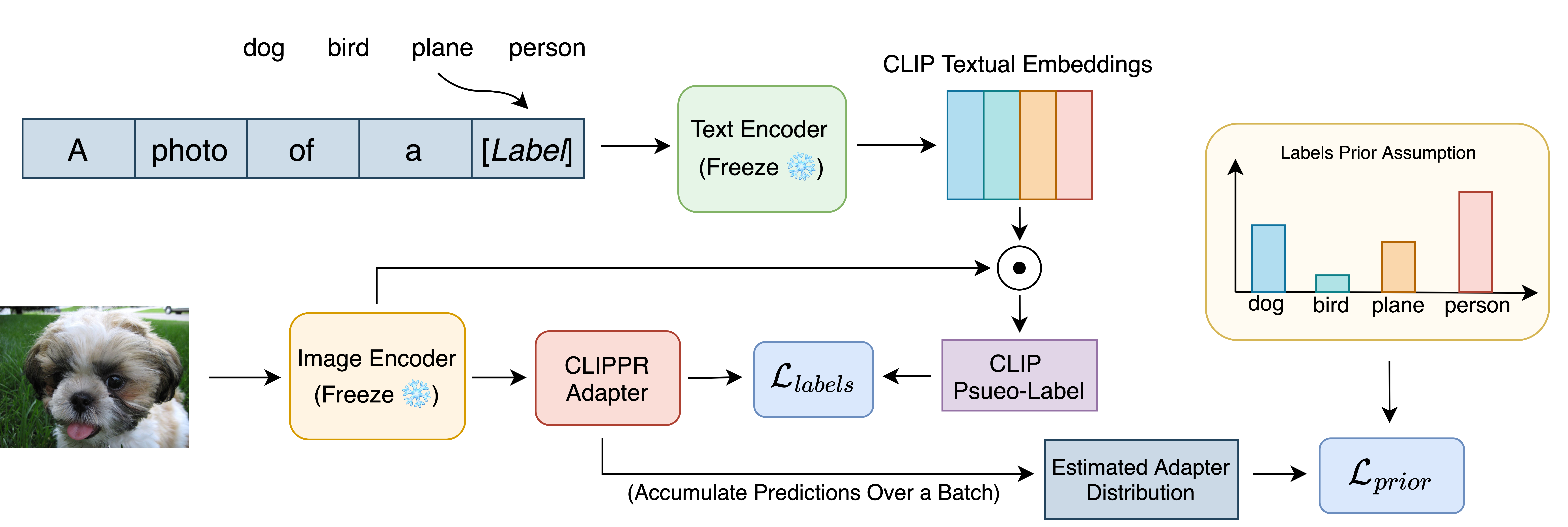}
    \caption{An illustration of our method. Our method trains an adapter module on top of a frozen \VL model image encoder. The adapter is trained with two competing objectives: (i) Predicting labels close to the original \VL model zero-shot predictions. (ii) Predicting a labels distribution similar to the given prior distribution. Together, these two objectives adapt the original zero-shot predictions to the distributional prior, resulting in better performance.}
    \label{fig:clippr_diag}
\end{figure*}

Indexing and grouping of image datasets require annotation of the relevant image attributes.
As manual annotation of large datasets is tedious and sometimes impossible, automatic methods are needed. Supervised machine learning methods are very effective when manual annotations are available, however these can only be provided for a handful of attributes. Standard methods are inadequate for the vast majority of image attributes, which lie at the long tail of the distribution, due to lack of manual labels.

Zero-shot learning \cite{socher2013zero,li2017learning} is a promising direction for addressing this need. The aim of such methods is to provide the means to create classifiers or regressors on-the-fly without requiring labeled examples. Recent advances in Vision \& Language models (\VL\!\!), have transformed our approach to zero-shot classification \cite{desai2021virtex,sariyildiz2020learning,zhang2020contrastive}, most notably CLIP \cite{radford2021learning}.

Training on large web-scale uncurated datasets allows CLIP to recognize a vast number of concepts and generalize to many domains (e.g., photos, drawings). This however causes a mismatch between CLIP's training distribution and that of the dataset we wish to label. Another issue is that the outputs of CLIP's text encoder are typically not as accurate as a linear classifiers trained on the visual features with annotated examples. Finally, while zero-shot classification has received much attention, the zero-shot \textit{regression} performance of \VL is often insufficient for accurate dataset annotation.

Our key idea is to adapt a zero-shot model using prior knowledge of the distribution of labels in the dataset.  While adaptation of classifiers to target label distributions have been studied before \cite{ji2019invariant}, the adaptation of \VL models poses unique challenges: (i) \VL models do not naturally yield an initial classification probability to be further calibrated. Instead, classification is carried using a cosine similarity metric, whose connection to a probability is indirect. (ii) Regression tasks may incorporate additional priors (e.g., continuity) not easily described by categorical probability. (iii) Zero-shot regression or classification methods might not have access to task-specific labelled training data at all, not even from a biased distribution. This prevents calculation of importance weights \cite{saerens2002adjusting,byrd2019effect}.

\begin{figure*}[t!]
    \centering
    \includegraphics[width=16cm]{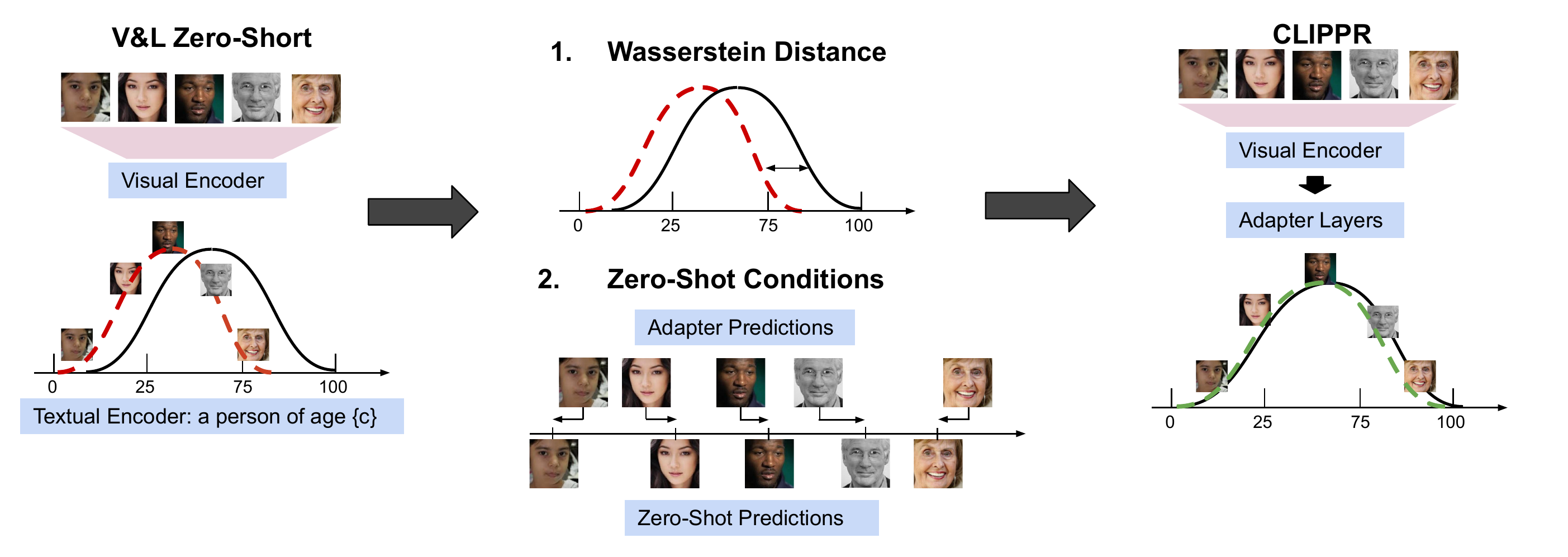}
    \caption{Our method adapts the zero-shot \VL for classification and regression using a prior on the distribution of labels. Left: The distribution of zero-shot labels (red dashed line) differs from that dataset prior (black line). Middle: We optimize our adapter to: (1) fit the distribution of predictions to the label prior (2) keep adapter predictions near their original values (predicted by the pre-trained \VL model). Right: The adapted predictions (green dashed line) follow the prior distribution resulting in a higher accuracy.}
    \label{fig:clippr}
\end{figure*}

To overcome these issues, we suggest to learn an adaptive module on top of the \VL model visual encoder using an unlabelled target set and a prior distribution on the labels. We rely \textit{only} on the \VL model's zero-shot predictions and an assumed target distribution. First, we choose an appropriate textual prompt based only on our prior for the unlabelled dataset. Second, we learn an adapter aimed at better fitting our zero-shot prediction to the given prior. Our adapter training objective is twofold: (i) Mapping each image sample close to its initial label assigned by CLIP (ii) Predicted labels should follow the provided distributional prior of the data. Specifically, we sample batches of the unlabelled data, and calculate the distance between their distribution and the prior distribution for these labels. We provide an illustration for our prior based approach in Fig. \ref{fig:clippr}. 

Finally, we demonstrate the advantage of using a prior distribution of the labels on zero-shot tasks. We target our method for zero-shot regression tasks, a mostly overlooked setting for zero-shot adapters, yet an important one for many concepts. We showcase our results on three regression benchmarks. We evaluate existing zero-shot prediction for this setting, and provide a significant improvement on three zero-shot regression tasks. The regression MAE (mean absolute error) on the UTK age regression dataset is improved from 7.99 to 5.73. Although originally targeted for regression tasks, we find that our method, with a simple adaptation, is also promising for classification tasks. We demonstrate an accuracy improvement of $2.83\%$ on the ImageNet \cite{deng2009imagenet} dataset. 

Our contributions are: (1) Using distributional priors for adapting pre-trained \VL models without any labelled data. (2) A loss designed for adapting pre-trained \VL models for zero-shot regression tasks. (3) Empirically demonstrating the benefits and robustness of the suggested adaptation method for zero-shot regression and classification.
(4) Proposing a novel method for selecting prompts for \VL zero-shot tasks given only an estimated distributional prior.

\section{Related Work}
\noindent \textbf{Automatic labelling of image datasets.} Training machine learning models for labelling natural images is one of the most studied tasks in computer vision \cite{lu2007survey}. While supervised methods have achieved very impressive results \cite{tan2019efficientnet}, the extensive need for supervision inspired many works aiming to learn with fewer labels \cite{chen2019closer,sohn2020fixmatch}. Another prominent line of works aims to use large unlabelled datasets to learn a strong visual representation, which can then be utilized for labelling downstream datasets with fewer supervised samples \cite{he2022masked,grill2020bootstrap,caron2021emerging}. Many clustering methods based on deep learning have been proposed for categorizing large dataset without any labels,  \cite{van2020scan,ji2019invariant}. Zero-shot models aim to categorize images without any labelled images from the target dataset, instead relying on labelling of specific attributes \cite{lampert2009learning,chao2016empirical,atzmon2019adaptive}; or on automatically collected online image captions \cite{mori1999image,radford2021learning}.

\noindent \textbf{\VL Models.} The ability to train \VL models on images with free language captions has been studied by many previous papers \cite{mori1999image,quattoni2007learning,joulin2016learning,sariyildiz2020learning,li2020mopro}. The impressive success of CLIP on zero-shot recognition and robustness to distribution shifts \cite{radford2021learning}, along with its many downstream applications \cite{agarwal2021evaluating}, inspired many methods aiming to further improve its performance. Recent models used advanced training techniques \cite{zhai2022lit,li2022masked,wang2021simvlm,wortsman2022robust,yao2021filip}, different architectures \cite{li2022masked}, or extended datasets \cite{jia2021scaling}. For simplicity, in this work we focus on adapting the CLIP model \cite{radford2021learning}, but the ideas we present are also applicable to more recent models.

\noindent  \textbf{\VL model adaptation for few shot classification.} The growing success of \VL models has inspired a line of works aiming to use a few annotated samples in order to boost their classification performance. Typically, these methods use an adapter layer learned on top of the \VL visual encoder \cite{gao2021clip,zhang2021tip}. Other methods used additional few-shot data to enhance the long-tailed recognition performance \cite{ma2021simple}. Another technique suggested, is to learn a dense prompt for the textual encoder, adapting the captions according to the few-shot supervision \cite{zhou2022learning,zhou2022conditional}.

\noindent  \textbf{Prior-shift adaptation.} In many computer vision tasks, the label distribution in the training data and test data may differ. A variety of prior-shift adaptation methods have been proposed to address this issue\cite{vsipka2022hitchhiker}. These methods can either use a given prior for the distribution of the labels in the test set\cite{saerens2002adjusting,sulc2019improving,vucetic2001classification}; or estimate this prior given an unlabelled dataset \cite{du2014semi,saerens2002adjusting,vucetic2001classification}. However, these methods require the distribution of labels in the training data, and therefore are not applicable to source-free adaptation of recent \VL method such as CLIP.

\section{CLIPPR: Adaptive Distribution Matching}
\label{sec:clippr}

\begin{table*}[t!]
\small
\begin{center}
\begin{tabular}{lccc}
\toprule
Prompt & MAE $\Downarrow$ (CLIP) & MAE $\Downarrow$ (CLIPPR) & $\mathcal{W}\Downarrow$\\
\midrule
'a photo from my $[label]$ birthday.' & 14.37 & 6.83 & 10.83 \\
'I was $[label]$ when they took this photo.' &  14.20 & 6.57 & 10.63 \\
'$[label]$ years old.' &  10.01 & 6.04 & 6.66 \\
'this person was born $[label]$ years ago.' & 10.15  & 6.41 & 5.33 \\
'a photo of a person, he is $[label]$ years old.' & 8.47 & 5.98 & 4.38 \\
'a person of age $[label]$.' & \textbf{7.96} &	\textbf{5.70} & \textbf{4.31} \\

\bottomrule
\end{tabular}
\end{center}
\caption{Zero-shot error (MAE) and Wasserstein distance ($\mathcal{W}$) from the prior of different prompts (UTK dataset).}
\label{tab:prompts}
\end{table*}

\subsection{Overview}

We propose  ``\textit{{CLIP} with {P}riors''} (CLIPPR), a new method for automatic dataset labeling with zero-shot learning. Our method does not assume any labelled samples for the target dataset. Instead, our method assumes as input a user estimate for the distribution of labels in the target dataset. It begins by choosing a prompt that yields zero-shot predictions that fit best our label distribution prior (Sec. \ref{sec:prompt_selection}). Then, we train an adapter module that modifies the predictions of the model to follow the label distribution prior, while also preserving the initial knowledge of the \VL model. A detailed illustration for our method is shown in Fig. \ref{fig:clippr_diag}.

We provide a method for optimizing this objective for regression, and a modified version for classification. In both cases, our method follows the same steps: (i) Adjusting the distribution of adapter predictions to fit the prior (Sec. \ref{sec:prior_distribution_matching}). Our technical approach differs for regression 
and classification. (ii) Requiring the predicted labels of the adapter to be similar to the original predictions of the pretrained \VL model (Sec. \ref{sec:conditioning_on_zs}). We empirically demonstrate the benefit of our method on multiple datasets in Sec.~\ref{sec:experiments}. 

\subsection{Preliminaries}
We rely on a pre-trained \VL model, with a text encoder $\CLIPT$ and an image encoder $\CLIPI$. 
We consider an unlabelled dataset of images $\mathcal{D}$. Our aim is to predict labels for these images, using a set of textual captions $\mathcal{C}$. Captions are identical except for a single word which describes each label. We denote the function that assigns image $x \in \mathcal{D}$ to its closest caption $c \in \mathcal{C}$ as $\mathcal{S}(\CLIPI(x), h^{\mathcal{T}}(\mathcal{C}))$.
For simplicity, we will denote $\mathcal{S}(x) \equiv \mathcal{S}(\CLIPI(x), h^{\mathcal{T}}(\mathcal{C}))$ going forward. Additionally, we assume an estimated prior distribution $\mathcal{P}(\mathcal{C})$, for the target labels $\mathcal{C}$. 
As we do not wish to learn an entirely new representation, but merely to modify it to match the distributional label, we keep both the text encoder $\CLIPT$ and visual encoder $\CLIPI$ frozen (Fig. \ref{fig:clippr_diag}). Instead of training a new representation, we train an adapter module $\mathcal{A}$, and set it to be relatively lightweight. The adapter aims to improve our prediction accuracy using the prior target distribution. The adapter $\mathcal{A}$, maps each image in the dataset $x \in \mathcal{D}$, to a prediction $y = \mathcal{A}\big(h^{\mathcal{I}}(x)\big)$ corresponding to one of the labels in $\mathcal{C}$. It terminates with a single neuron for regression tasks, or a SoftMax layer for classification.

\subsection{Label Distribution Prior}

\VL models are trained on massive, self-supervised web datasets. While this is necessary for them to recognize many concepts, it causes a mismatch between their training dataset and our target datasets. 
For example, if the vast majority of people in the images are from a certain age group, it may distort the way a \VL model perceives this attribute in new images encountered during inference. We therefore require the outputs of our adapter $\mathcal{A}$ to obey the distributional prior $\mathcal{P(C)}$:

\begin{equation} 
 \mathcal{A}\big(h^{\mathcal{I}}(x)\big)_{ x \sim \mathcal{D}} \sim \mathcal{P}(\mathcal{C})
\label{eq:distrbution_sim}
\end{equation}

\subsection{Automatic Prompt Selection}
\label{sec:prompt_selection}

\begin{figure*}[htp]
    \centering
    \includegraphics[width=14cm]{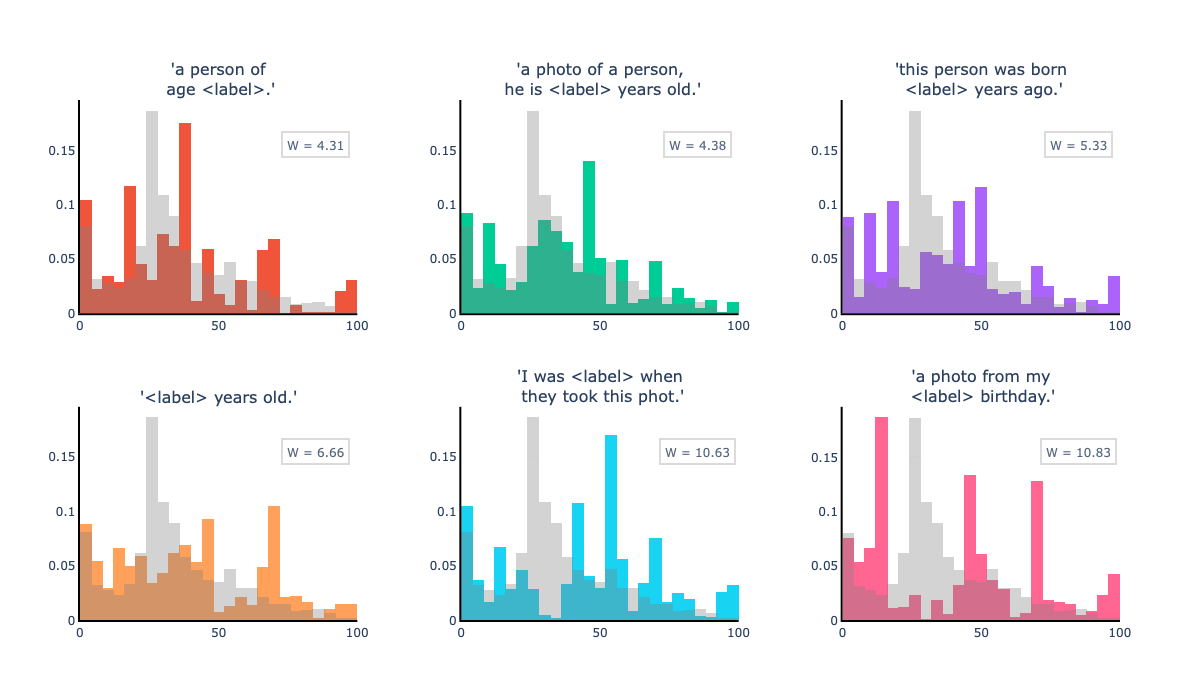}
    \caption{The proximity of zero-shot predictions on a dataset to the prior distribution of labels predicts successful prompts for zero-shot labeling. In each of the plots we present: i) The age distribution predicted on the UTK dataset by CLIP using the prompt in its title. ii) The Wasserstein between it and our prior distribution ($W$, floating box). 
    As shown in Tab. \ref{tab:prompts} ,the smaller the distance between the distributions, the better the prompt predicts the ground-truth labels (gray histogram in the background). }
    \label{fig:prompts}
\end{figure*}

We use a novel method for selecting effective task-specific prompts. Provided a set of possible prompts, we extract CLIP zero-shot predictions for all images in the dataset, for each prompt. We then compare the different prompts by the Wasserstein distance between: (i) The predicted distribution of labels for the entire dataset using the prompt. (ii) The prior distribution of labels (see also Sec. \ref{sec:prior_distribution_matching}). This criterion is highly predictive of the zero-shot prediction accuracy, both for CLIP and for our method. We visually present in Fig. \ref{fig:prompts} the predicted distribution of labels on the UTK dataset. Their relation to zero-shot performance is presented in Tab.~\ref{tab:prompts}. 

\subsection{Prior Distribution Matching}
\label{sec:prior_distribution_matching}

While selecting an appropriate prompt can help us match the predicted label distribution with the prior, it is usually insufficient. Therefore, we train an adapter module to better match the distributions. We use different training procedures for regression and classification.

\subsubsection{Regression}
\label{sec:prior_distribution_matching_reg}
Enforcing Eq.~\ref{eq:distrbution_sim} for regression requires measuring the distance between distributions of continuous variables. For example, given a dataset of human faces, we would wish the distribution of predicted ages to match a prior distribution of ages in the population (Fig. \ref{fig:clippr} ). Our method first samples a set of images and a set of labels $Y_{\mathcal{P}(\mathcal{C})}$ from the prior $\mathcal{P}(\mathcal{C})$. Then, we predict the numeric values of the sampled images $Y_{\mathcal{A}}$ using the adapter module. Finally, we minimize the empirical distributional distance between the samples from the adapter $Y_{\mathcal{A}}$ and the samples from our prior $Y_{\mathcal{P}(\mathcal{C})}$. 

To measure the distances between the two empirical distributions, we minimize the Wasserstein distance \cite{vallender1974calculation} ($\mathcal{W}$) between the two sets of labels. 
Intuitively, this minimizes the average difference between a matching pair of prior and predicted labels, weighted by their probabilities. 
It is a differentiable metric, and has a particularly simple form for one-dimensional labels. For our setting, this translates to first sorting both $Y_{\mathcal{A}}$ and $Y_{\mathcal{P}(\mathcal{C})}$ in order of increasing size. We denote the $i$th smallest labels as  $Sort(Y_{\mathcal{A}})[i]$ and $Sort(Y_{\mathcal{P}(\mathcal{C})})[i]$. The Wasserstein distance is then given by the following expression:

\begin{equation}
     W(Y_{\mathcal{A}},Y_{\mathcal{P}(\mathcal{C})}) = \sum_{i=1}^{B} | Sort(Y_{\mathcal{A}})[i] - Sort(Y_{\mathcal{P}(\mathcal{C})})[i] |_1
    \label{eq:batch_wasserstein}
\end{equation}

\noindent Where $B$ denotes batch size. \noindent We can therefore denote our optimization objective as:

\begin{equation}
    \mathcal{L}_{prior\_reg}\big( \mathcal{A}, \mathcal{P}(\mathcal{C})\big) = W(Y_{\mathcal{A}},Y_{\mathcal{P}(\mathcal{C})})
     \label{eq:loss_wasserstein}
\end{equation}

where $Y_{\mathcal{P}(\mathcal{C})} \sim \mathcal{P}(\mathcal{C})$ and $Y_{\mathcal{A}} = \mathcal{A}\big(h^{\mathcal{I}}(x)\big)_{x \sim  D}$ \\

\noindent In expectation (and over large enough batch sizes $|B|$), this distance is minimized when our adapter predicts the same label distribution as the prior, on the entire dataset.

\subsubsection{Classification}
\label{sec:prior_distribution_matching_clf}
Our prior distribution $\mathcal{P}(\mathcal{C})$ for classification is a multinomial distribution. Similarly to our approach for regression, we minimize an empirical distance between batches of our predicted labels and the prior distribution.  
We first average the probability of each label across the entire batch, to gain an estimate for the empirical predicted labels distribution $Y_\mathcal{A}$.

\begin{equation}
   {Y}_\mathcal{A}[c] = \frac{1}{|B|} \sum_{x \in B} P_{\mathcal{A}}(c|\CLIPI(x))
   \label{eq:cls_dist}
\end{equation}

\noindent where $B \subset D $ is a batch of images, and $P_{\mathcal{A}}(c|\CLIPI(x))$ is the probability for a class $c \in \mathcal{C}$ given by the adapter for an image $x \in B$. 

We then minimize the $KL$-Divergence between the prior distribution and the predicted label distribution.

\begin{equation}
    \mathcal{L}_{prior\_clf}\big( \mathcal{A}, \mathcal{P}(\mathcal{C})\big) = D_{KL}(Y_\mathcal{A},\mathcal{P}(\mathcal{C}))
     \label{eq:loss_dkl}
\end{equation}

\subsection{Conditioning on Zero-Shot Predictions}
\label{sec:conditioning_on_zs}
Having the distribution of predicted labels match the prior is necessary but clearly insufficient. In the extreme case, we may have an adapter whose predictions are distributed exactly as $\mathcal{P}(\mathcal{C})$ over the entire dataset, while not a single image is correctly labelled.
To encourage our predictions to be semantically meaningful, we would like them to retain most of the knowledge embedded in the original \VL encoders. We therefore require the predictions of the adapter $\mathcal{A}$ to remain similar to the labels originally predicted  by zero-shot of the pre-trained \VL model. We begin with a hard-assignment of an initial label to each of the (unlabelled) samples in the dataset, using the chosen prompt:

\begin{equation} 
\hat{y_i} = \mathcal{S}(x_i)
\end{equation}

\noindent We train the adapter layer $\mathcal{A}$ to output this prediction. For regression, we use an $\ell_1$ loss between the adapter and original predictions:
    \begin{equation} 
        \mathcal{L}_{labels\_reg}(x_i, \hat{y_i}) = \ell_1\big(\mathcal{A}(h^{\mathcal{I}}(x_i)), \hat{y_i}\big)
    \end{equation}
    
For classification, we use the $CrossEntropy$ objective to optimize prediction accuracy for the desired label:

    \begin{equation} 
        \mathcal{L}_{labels\_clf}(x_i, \hat{y_i}) = CE\big(\mathcal{A}(h^{\mathcal{I}}(x_i)), \hat{y_i}\big)
    \end{equation}

\subsection{CLIPPR: CLIP with Priors}

Our full training objective combines the label prior and zero-shot conditioning loss terms:

\begin{equation}
    \mathcal{L}_{CLIPPR} = \mathcal{L}_{prior} + \alpha \cdot \mathcal{L}_{labels}
    \label{eq:clipper_combined}
\end{equation}

\noindent where $\alpha$ is a hyper-parameter for weighting the two losses. Implementation details can be found in Sec. \ref{sec:imp_details} and \ref{appendix:imp_details}.

\section{Experiments}   
\label{sec:experiments}
We evaluate our method on both classification and regression tasks. Our zero-shot regression results are presented first, followed by our zero-shot classification results. 

\subsection{Baselines}

We compare our method, CLIPPR, with the following baselines:

\textit{CLIP}: We use CLIP in its raw pre-trained form, with the set of textual labels provided with the dataset. For each image we choose to assign the label $c \in \mathcal{C}$ that is most similar in embedded space to that image embedding. 

\textit{Supervised}: Here, we use the labels of the train set to train the adapter layers directly. We use a $CrossEntropy$ loss or $\ell_1$ loss for our classification or regression experiment, respectively. This evaluation uses training labels, which are not part of our setting. Rather, we use it as an estimate of the upper bound for the performance an adaptation method can achieve.

\subsection{Implementation Details}
\label{sec:imp_details}

\noindent\textbf{Prior distributions}: 
\noindent For the regression task on the UTK dataset, we used a Gaussian distribution:
$\mathcal{P}(\mathcal{C}) = \frac{1}{\sigma\sqrt{2\pi}} e^{-\frac{1}{2} (\frac{x-\mu}{\sigma})^2}$,
where the average label value $\mu$ and standard deviation $\sigma$ were determined by the empirical distribution in the dataset. We present the target and prior distributions for this dataset in Fig. \ref{fig:robust}. For the Stanford Cars dataset, the regression values come from 16 categories, and we used an appropriate prior distribution according to their abundances in the dataset. As the Adience dataset is typically labelled by a set of ranges, we apply the same protocol for it as for the Stanford Cars dataset. For all classification datasets, we simply assumed an equal probability for each of the possible captions as our prior.

\noindent\textbf{Data splits:} Although our method improves the zero-shot prediction on its unlabelled training data, we report our results on the test set for compatibility with the compared methods. 

\noindent\textbf{Combination of loss terms:} We simply used an equal weight combination ($\alpha = 1$, Eq. \ref{eq:clipper_combined}) in all our experiments. We discuss this choice in Sec. \ref{sec:alpha}.

\noindent\textbf{Prompting}: We use a single prompt for each dataset without using any labels or training procedure. For regression tasks, the list of captions contained all the relevant numerical values (e.g., any age between 0 and 100). For classification tasks, we use the provided list of class names, following previous works \cite{radford2021learning}. Full details regarding the captions and prompts can be found in the supplementary material. We note that our zero-shot results differ from the original CLIP paper \cite{radford2021learning}. All exact prompts and captions, along further implementation details can be found in the supplementary material.

\subsection{Regression}   
We test the effectiveness of our approach on different regression datasets. The selected datasets are all commonly used and have at least one annotated numeric attribute. We score the compared methods using mean absolute error accuracy on the test set:

\textit{UTKFace Large Scale Face Dataset \cite{zhifei2017cvpr}}: This dataset consists of over $20K$ images of human faces, annotated in terms of age, gender and ethnicity. We aim to predict the age of the person presented in each image as a number between $0$ and $116$.

\textit{Stanford Cars \cite{KrauseStarkDengFei2013}}: This dataset contains over $16K$ car images with different backgrounds. Our regression task here is to predict the production year of the car in each image (between 1993 and 2012). As the dataset is very biased in terms of production years, we modify it slightly by using only half of the samples from 2012 (which originally are $60\%$ of the data), so it is more balanced in terms of our target attribute. 

\textit{Adience \cite{eidinger2014age}}: The Adience dataset contains over $26K$ images of people gathered from over $2.2K$ different subjects. Our regression task on this dataset is to determine the age of the person present in each image. When the images in the dataset are labelled with a range of ages, we use the average year as the label.


\subsubsection{Results}

Our method outperformed the original CLIP model on all $3$ benchmarks (Tab. \ref{tab:regression}). Evaluating human age predictions on the UTK and Adience we are able to approach the \textit{supervised} upper limit for adaptation methods. For the Stanford Cars dataset, we again find a significant improvement, reducing the gap from supervised performance in half.

\begin{table}[h!]
\small

\begin{center}
\begin{tabular}{lccc}
& \textbf{UTK} & \textbf{StanfordCars} & \textbf{Adience}	\\
\toprule
CLIP Zero-Shot & 7.99 & 2.72 & 7.47 \\
CLIPPR (Ours) & \textbf{5.73} {\textbf{\scriptsize\textcolor{limegreen}{(-2.26)}}} & \textbf{2.17} {\textbf{\scriptsize\textcolor{limegreen}{(-0.52)}}} & \textbf{5.92} {\textbf{\scriptsize\textcolor{limegreen}{(-1.57)}}} \\
\midrule
CLIP Supervised & 4.21 & 1.671  & 4.68 \\	
\bottomrule
\end{tabular}
\end{center}
\caption{Zero-shot regression error $(\Downarrow)$ of different adaptation methods (Mean Absolute Error). Performance improvement from CLIP is shown in {\textbf{\textcolor{limegreen}{green}}}.}
\label{tab:regression}
\end{table}

\begin{figure*}[h!]
    \centering
    \includegraphics[width=15cm]{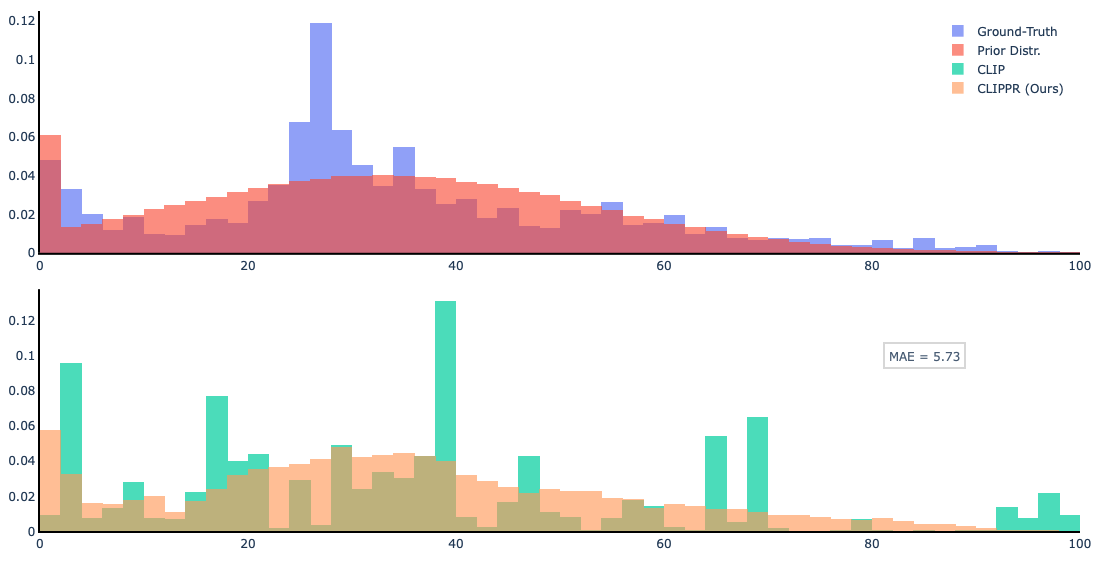}
    \caption{Our methods fits the predicted labels distribution to the prior improving the regression accuracy, even when the prior deviates from the real distribution. Top: The ground truth distribution of ages in the UTK dataset (blue) and the prior distribution we used for it (red). Bottom: The zero-shot age prediction by CLIP (green) and by our method, CLIPPR (orange).}
    \label{fig:robust}
\end{figure*}

\subsection{Classification}   

We demonstrate the ability of our method to improve the labelling of image datasets using two of the most commonly used image classification datasets:

\textit{Cifar10 \cite{krizhevsky2009learning}:} The Cifar10 dataset contains $32 \times 32$ resolution images of 10 coarse-grained object categories. The image classes describe different animal species and vehicle types. 

\textit{ImageNet \cite{deng2009imagenet}:} A large-scale dataset containing 1 million images from $1000$ object category classes taken from the WordNet dataset. It has become central to computer vision and deep learning research due to the large number of images and classes.

\subsubsection{Results}

While the main objective of our method is to improve zero-shot regressions, we find that it also performs well for classification (Tab. \ref{tab:classification}). On ImageNet, we see a significant improvement of almost $3\%$ in accuracy. In CIFAR10, our method halved the gap in classification error between the performance of CLIP and the fully supervised baseline, reaching a classification accuracy of $90\%$. Our method showed significantly better results than the original predictions of CLIP, on both datasets.

\begin{table}[h!]
\small
\begin{center}
\begin{tabular}{lcc}
& \textbf{ImageNet} & \textbf{CIFAR10} \\
\toprule
CLIP Zero-Shot & 57.59 & 85.17 \\
CLIPPR (Ours) & \textbf{60.42} {\textbf{\scriptsize\textcolor{limegreen}{(+2.83)}}} & \textbf{89.74} {\textbf{\scriptsize\textcolor{limegreen}{(+4.57)}}} \\
\midrule
CLIP Supervised & 68.04 & 96.14 \\	
\bottomrule
\end{tabular}
\end{center}
\caption{Zero-shot classification accuracies of different adaptation methods (Accuracy \%). Performance improvement w.r.t. CLIP is shown in {\textbf{\textcolor{limegreen}{green}}}.}
\label{tab:classification}
\end{table}

\subsection{Ablation Studies}

We conducted two ablation studies for our method. We evaluated: (i) our method robustness to inaccuracies in the given prior (ii) various weightings between the objectives.

\subsubsection{Robustness to Prior Inaccuracies}

Our method relies on an estimate of the prior distribution of labels which may be inaccurate. Mismatches between the prior distribution we use and the actual distribution of ground truth labels, may impair our method ability to improve zero-shot prediction capabilities. We therefore desire our method to be robust to prior distribution inaccuracies. First, we can see that our method achieves a significant performance improvement over CLIP even for a highly inaccurate Gaussian prior used for the UTK dataset (see Fig. \ref{fig:robust}); achieving MAE reduction from $7.99$ to $5.73$.

We also analyze two additional types of inaccuracies: (i)~A Gaussian prior with a biased mean (ii) A Gaussian prior with a biased variance. Results shown in Tab. \ref{tab:robustness} demonstrate some degree of robustness of our method in both aspects. Small changes in either shift or scale, merely affect the results. Our method experiences performance reductions when the prior is highly inaccurate, e.g. a mean of $>45$ ($140\%$ of true value) or if we use a standard deviation of $>30$ ($150\%$ of true value). When the prior has a very weak alignment with the true target distribution, our method will experience large performance drops.

\begin{table}[h!]
\small
\begin{center}
\begin{tabular}{lclc}
\multicolumn{2}{c}{\normalsize\textbf{Shift}} & \multicolumn{2}{c}{\normalsize\textbf{Scale}} \\
\toprule
$\mathcal{P(C)}$ & MAE $\Downarrow$ & $\mathcal{P(C)}$ & MAE $\Downarrow$\\
\midrule
$\mathcal{N}(45,\,20^{2})$ & 9.25 {\textbf{\scriptsize\textcolor{ered}{(+3.52)}}} & $\mathcal{N}(30,\,35^{2})$ & 8.26 {\textbf{\scriptsize\textcolor{ered}{(+2.51)}}}  \\	

$\mathcal{N}(40,\,20^{2})$ & 7.14 {\textbf{\scriptsize\textcolor{ered}{(+1.41)}}} & $\mathcal{N}(30,\,30^{2})$ & 7.28 {\textbf{\scriptsize\textcolor{ered}{(+1.53)}}}  \\	

$\mathcal{N}(25,\,20^{2})$ & 6.49 {\textbf{\scriptsize\textcolor{ered}{(+0.76)}}} & $\mathcal{N}(30,\,10^{2})$ & 6.90 {\textbf{\scriptsize\textcolor{ered}{(+1.15)}}}  \\	

$\mathcal{N}(35,\,20^{2})$ & 5.91 {\textbf{\scriptsize\textcolor{gray}{(+0.18)}}} & $\mathcal{N}(30,\,25^{2})$ & 6.36 {\textbf{\scriptsize\textcolor{ered}{(+0.61)}}}  \\	

$\mathcal{N}(30,\,20^{2})$ & 5.75 {\textbf{\scriptsize\textcolor{gray}{(+0.02)}}} & $\mathcal{N}(30,\,15^{2})$ & 5.75 {\textbf{\scriptsize\textcolor{gray}{(+0.00)}}}  \\	
\midrule
\multicolumn{2}{c}{$\mathcal{N}(33,\,20^{2})$  \textbf{5.73}} & \multicolumn{2}{c}{$\mathcal{N}(30,\,20^{2})$  \textbf{5.75}}  \\	
\bottomrule
\end{tabular}
\end{center}
\caption{Robustness to inaccuracies in the prior distribution  (Accuracy \%). Large performance gaps are shown in {\textbf{\textcolor{ered}{red}}}, while small performance gaps are shown in {\textbf{\textcolor{gray}{gray}}}.}
\label{tab:robustness}
\end{table}

\subsubsection{Weighting the Two Objectives}
\label{sec:alpha}
We perform an ablation test to study our robustness to different weightings of our two optimization losses ($\alpha$ from Eq. \ref{eq:clipper_combined}). Hyper-parameter selection for each dataset is not possible in practice, as we operate in the unsupervised setting. While our method works consistently with the same hyper-parameter for all experiments, we also present results with different values of $\alpha$ across three datasets in Tab. \ref{tab:alpha_ablation}. We find that an equal weighting of the two objectives yields optimal or near optimal performance, across all three datasets. Note, that for all three datasets any choice of $\alpha \in {0.3, 1, 3}$ always results in a substantial improvement over CLIP zero-shot performance, as mentioned in Tab. \ref{tab:regression} \& \ref{tab:classification}.

\begin{table}[h!]
\small
\begin{center}
\begin{tabular}{lccc}
& \textbf{UTK}$(\Downarrow)$ & \textbf{StanfordCars}$(\Downarrow)$ & \textbf{ImageNet}$(\Uparrow)$ \\
\toprule
$\alpha=10$ & 10.10 & 4.69 & 59.86 \\
$\alpha=0.3$ & 6.02 & 2.36 & 59.71 \\
$\alpha=3$ & 6.16 & 2.27 & \textbf{60.58} \\	
\midrule
$\alpha=1$ & \textbf{5.73} {\textbf{\scriptsize\textcolor{limegreen}{(-0.29)}}} & \textbf{2.17} {\textbf{\scriptsize\textcolor{limegreen}{(-0.1)}}} & 60.42 {\textbf{\scriptsize\textcolor{ered}{(-0.16)}}} \\	
\bottomrule
\end{tabular}
\end{center}
\caption{Ablation of the $\alpha$-weighing of the two objectives. Regression error is measured by the Mean Absolute Error $(\Downarrow)$, while classification is measured by accuracy $(\Uparrow)$.}
\label{tab:alpha_ablation}
\end{table}

\section{Discussion} 

\textbf{Comparison with few-shot adapters.} Our method does not require having any annotated images, but does require having a reasonable guess of the label distribution. Conversely, much research \cite{zhou2022conditional,zhang2021tip} was done on few-shot learning: where a few labeled examples from each label are provided in addition to the pre-trained \VL model. We compare our method on the UTK dataset (without labelled samples) to Tip-Adapter \cite{zhang2021tip}, a state-of-the-art \VL few-shot method. We achieved better results than Tip-Adapter with $32$ shots per age group. This shows the promise of our approach for zero-shot regression tasks.

\begin{table}[h!]
\small
\begin{center}
\begin{tabular}{lc}
& \textbf{MAE} $(\Downarrow)$ \\
\toprule
Tip-adapter (1-shot) & 7.19 {\textbf{\scriptsize\textcolor{ered}{(+1.46)}}} \\
Tip-adapter (2-shots) & 7.58 {\textbf{\scriptsize\textcolor{ered}{(+1.85)}}} \\
Tip-adapter (4-shots) & 6.43 {\textbf{\scriptsize\textcolor{ered}{(+0.70)}}} \\
Tip-adapter (8-shots) & 6.30 {\textbf{\scriptsize\textcolor{ered}{(+0.57)}}} \\
Tip-adapter (16-shots) & 6.12 {\textbf{\scriptsize\textcolor{ered}{(+0.39)}}} \\
Tip-adapter (32-shots) & 5.94 {\textbf{\scriptsize\textcolor{ered}{(+0.21)}}} \\
\midrule
CLIPPR (Ours) & \textbf{5.73} \\
\bottomrule
\end{tabular}
\end{center}
\caption{Few-shot regression performance (Mean Absolute Error $\Downarrow$) of Tip-Adapter compared to our method, on the UTK dataset.}
\label{tab:few_shot}
\end{table}

\textbf{Multivariate regression.} In this work, we dealt with regression tasks where the target labels are univariate. We believe our method can be extended to the setting where the output is a multivariate variable. This can be done by replacing our distribution matching criteria to the sliced Wasserstein distance \cite{kolouri2019generalized}. The sliced Wasserstein distance first projects both the source and target distributions along a set of random axes. It then computes the 1-D Wasserstein distance as presented in Sec.~\ref{sec:clippr} for each projection. Finally, it averages the distances from all the projections. 
As estimating a prior for the multivariate-output setting might be a non-trivial task, we leave this investigation for future work.

\noindent \textbf{Distribution matching in the feature space.} Our method aligns the predicted distribution with a prior by matching them directly. We believe that aligning the distribution earlier, in the \VL model embedding space, might help converging to a better solution. However, it is not obvious how to estimate the prior in the embedding space. We encourage future research to explore this direction.

\section{Limitations}   

\textbf{Failure cases of CLIP.} We found a few examples where the zero-shot predictions by CLIP were too inaccurate for our method. For example, when evaluated on a regression task of estimating the production year of a mobile phone, we found that CLIP's predictions were no better than a random guess. In such a setting, our method's best guess was to randomly pick labels according to the prior distribution. Our method relies on CLIP's zero-shot labels, so when CLIP fails completely, our method also fails. 

\textbf{Captions where CLIP is not applicable.} CLIP also fails with concepts that are not easily described within the context of a caption. For example, while it is possible to describe the specific angle of an object in a few sentences, prompting a \VL for it is difficult. 
Therefore, CLIP's text encoder is unlikely to describe it well. As above, a complete failure of CLIP would also indicate failure of our method.

\textbf{Estimating a prior.} Our method requires estimating a prior for the distribution of labels. Although we show our method is relatively robust to the exact choice, obtaining a reasonable prior might be difficult in certain scenarios. We therefore include it as a limitation of our method.

\section{Conclusion}   
We presented a new approach for enhancing zero-shot performance of \VL models using a prior on the distribution of labels, but no annotated images. We evaluated it on both regression and classification tasks. Our method achieved a significant improvement over the pretrained zero-shot performance of CLIP. 

\section{Acknowledgment}   

This work was partially supported by the Center for Interdisciplinary Data Science Research (CIDR) and the Israel Council for Higher Education.

{\small
\bibliographystyle{ieee_fullname}
\bibliography{egbib}
}

\clearpage

\include{ArXiv_suppl}

\end{document}

%% file: ArXiv_suppl.tex
{\LARGE{Appendix}}

\section{Implementation Details}
\label{appendix:imp_details}

\noindent \textbf{Captions \& Prompts.} We used the following prompts and captions:

\begin{itemize}

    \item \textbf{UTK.} \\ 
    \textit{prompt:} `a person of age $[label]$.' \\ 
    \textit{captions:} We use all ages found in the dataset between $1-100$. We label all images of people aged $100$ and above as $100$, due to lack of data in those ages.
    
    \item \textbf{Stanford Cars.} \\ 
    \textit{prompt:} `a car from $[label]$.' \\ 
    \textit{captions:} The production years of all vehicles featured in the dataset. Specifically: 2012-2006, 2002-1997, 1994, 1993, and 1991.

    \item \textbf{Adience.} \\ 
    \textit{prompt:} `age $[label]$.' \\ 
    \textit{captions:} We use all ages in the dataset. For the images in the dataset labelled by a range of ages, we take the range average. We are left with the following ages as our group of captions: 1,  2,  3,  5, 10, 13, 16, 18, 22, 23, 28, 29, 30, 32, 34, 35, 36, 40, 42, 43, 45, 46, 50, 55, 56, 57, 58, and 80.

    \item \textbf{CIFAR10.} \\ 
    \textit{prompt:} `a photo of a $[label]$.' \\ 
    \textit{captions:} We simply use the class names as our set of captions: airplane, automobile, bird, cat, deer, dog, frog, horse, ship, and truck.

    \item \textbf{ImageNet.} \\ 
    \textit{prompt:} `a photo of a $[label]$.' \\ 
    \textit{captions:} We use a simplification of the original ImageNet class names\footnote{\href{https://github.com/anishathalye/imagenet-simple-labels}{https://github.com/anishathalye/imagenet-simple-labels}}.

\end{itemize}

\noindent \textbf{Optimization Parameters.} All of our experiments were performed using a ViT-B/32 backbone CLIP visual encoder. For the experiments which require training (all except CLIP Zero-shot), we trained an adapter over CLIP image head for $70$ epochs, with a weight decay regularization of $0.0001$, and an initial learning rate of $0.001$; decayed by $0.3$ every $10$ epochs. An exception to that is the ImageNet dataset where, due to its large size, we trained the adapter for $7$ epochs, decaying the learning rate by $0.3$ every epoch. We used a fixed batch size of $128$, as well as the Adam \cite{} optimizer with moments $\beta_1 = 0.9, \beta_2 = 0.999$, for all of our trained experiments.

\subsection{Batch Accumulation for Classification Tasks}

When training our method on classification datasets, we relied on having large enough batches which approximate the predicted distribution of the labels. This is not true when training with a batch size of $128$ and having $1000$ ImageNet classes. Due to limited resources, we are not able to increase the batch size any further. For this reason, in classification tasks only, we perform a gradient update every several batches instead of every single one. Therefore, for each gradient update we: (i) aggregate the $\mathcal{L}_{labels}$ term from Eq. $8$, over several batches. (ii) use predictions from multiple batches to estimate the adapter's predicted label distribution ($Y_{\mathcal{A}}$ from Eq. $4$). 

The CIFAR10 dataset has only $10$ classes, therefore we perform an update every $4$ batches, effectively equal to sampling a batch of $512$ images. The ImageNet dataset however, has $1000$ different classes. In order to estimate the predicted label distribution accurately, we have to aggregate our predictions over many batches. Specifically, we perform a gradient update every $40$ batches, which is equivalent to a batch size of $5120$ samples. 

In the ImageNet dataset, the total number of gradient updates is much smaller. This created issues for the convergence of the network. While this could be solved by more epochs, we again prefer a more efficient solution due to our limited resources. We choose to better initialize our adapter (in an unsupervised manner). This could be done once for all experiments on this dataset. We initialize our adapter by training it only with the $\mathcal{L}_{labels}$ objective, using CLIP zero-shot predictions as labels, for $3$ epochs. We find this initialization effective for our method.